\documentclass[aos,preprint]{imsart}

\usepackage{amsmath}
\usepackage{amssymb}
\usepackage{graphicx}
\usepackage{color}
\usepackage{booktabs}
\usepackage{multirow}
\usepackage{booktabs}
\usepackage{subfig}
\usepackage{microtype}
\usepackage[round]{natbib}
\usepackage{url}
\usepackage{afterpage}

\setattribute{journal}{name}{}
\startlocaldefs
\usepackage{macros}
\urldef\rpaurl\url{http://www.cs.toronto.edu/~rpa}
\endlocaldefs

\begin{document}
\begin{frontmatter}
  \title{Incorporating Side Information in Probabilistic Matrix
    Factorization with Gaussian Processes}
  \runtitle{Incorporating Side Information in Matrix Factorization}
  \begin{aug}
    
    \author{\fnms{Ryan P.}  \snm{Adams}\corref{}\thanksref{t2}\ead[label=e1]{rpa@cs.toronto.edu}},
    \author{\fnms{George E.} \snm{Dahl}\ead[label=e2]{gdahl@cs.toronto.edu}}
    \and
    \author{\fnms{Iain}  \snm{Murray}%
      \ead[label=e3]{murray@cs.toronto.edu}}
    \thankstext{t2}{\rpaurl}

     \runauthor{R.P.\ Adams et al.}
     \affiliation{University of Toronto}

  \address{Department of Computer Science\\
    University of Toronto\\
    10 King's College Road\\
    Toronto, Ontario M5S 3G4, Canada\\
    \printead{e1,e2,e3}}
  \end{aug}

\begin{abstract}
  Probabilistic matrix factorization (PMF) is a powerful method for
  modeling data associated with pairwise relationships, finding use in
  collaborative filtering, computational biology, and document analysis,
  among other areas.  In many domains, there is additional information that
  can assist in prediction.  For example, when modeling movie ratings, we
  might know when the rating occurred, where the user lives, or what actors
  appear in the movie.  It is difficult, however, to incorporate this side
  information into the PMF model.  We propose a framework for incorporating
  side information by coupling together multiple PMF problems via Gaussian
  process priors.  We replace scalar latent features with functions that
  vary over the space of side information.  The GP priors on these
  functions require them to vary smoothly and share information.  We
  successfully use this new method to predict the scores of professional
  basketball games, where side information about the venue and date of the
  game are relevant for the outcome.
\end{abstract}
\end{frontmatter}

\section{Introduction}
Many data that we wish to analyze are best modeled as the result of a
pairwise interaction.  The pair in question might describe an interaction
between items from different sets, as in collaborative filtering, or might
describe an interaction between items from the same set, as in social
network link prediction.  The salient feature of these \textit{dyadic} data
modeling tasks is that the observations are the result of interactions: in
the popular Netflix prize example, one is given user/movie pairs with
associated ratings and must predict the ratings of unseen pairs.  Other
examples of this sort of relational data include biological pathway
analysis, document modeling, and transportation route discovery.

One approach to relational data treats the observations as a matrix and
then uses a probabilistic low-rank approximation to the matrix to discover
structure in the data.  This approach was pioneered by
\citet{hofmann-1999a} to model word co-occurrences in text data.  These
\textit{probabilistic matrix factorization} (PMF) models have generated a
great deal of interest as powerful methods for modeling dyadic data.  See
\citet{srebro-2004a} for a discussion of approaches to machine learning
based on matrix factorization and \citet{salakhutdinov-mnih-2008b} for a
current view on applying PMF in practice.

One difficulty with the PMF model is that there are often more data
available about the observations than simply the identities of the
participants.  Often the interaction itself will have additional labels
that are relevant to the prediction task.  In collaborative filtering, for
example, the date of a rating is known to be important \citep{koren-2009a}.
Incorporating this side information directly as part of the low-rank
feature model, however, limits the effect to only simple, linear
interactions.  In this paper we present a generalization of probabilistic
matrix factorization that replaces scalar latent features with functions
whose inputs are the side information.  By placing Gaussian process priors
on these latent functions, we achieve a flexible nonparametric Bayesian
model that incorporates side information by introducing dependencies
between PMF problems.

\section{The Dependent PMF Model}
In this section we present the \textit{dependent probabilistic matrix
  factorization} (DPMF) model.  The objective of DPMF is to tie together
several related probabilistic matrix factorization problems and exploit
side information by incorporating it into the latent features.  We
introduce the standard PMF model first and then show how it can be extended
to include this side information.

\subsection{Probabilistic Matrix Factorization}
\label{sec:pmf}
In the typical probabilistic matrix factorization framework, we have two
sets,~$\mcM$ and~$\mcN$, of sizes~$M$ and~$N$.  For a collaborative
filtering application,~$\mcM$ might be a set of films and~$\mcN$ might be a
set of users.  $\mcM$~and~$\mcN$ may also be the same sets, as in the
basketball application we explore later in this paper.  In general, we are
interested in the outcomes of interactions between members of these two
sets.  Again, in the collaborative filtering case, the interaction might be
a rating of film~$m$ by user~$n$.  In our basketball application, the
observations are the scores of a game between teams~$m$ and~$n$.  Our goal
is to use the observed interactions to predict unobserved interactions.
This can be viewed as a \textit{matrix completion} task: we have
an~${M\stimes N}$ matrix~$\bZ$ in which only some entries are observed and
must predict some of the unavailable entries.

One approach is to use a generative model for~$\bZ$.  If this model
describes useful interactions between the rows and columns of~$\bZ$, then
inference can provide predictions of the unobserved entries.  A typical
formulation draws~$\bZ$ from a distribution that is parameterized by an
unobserved matrix~$\bY$.  This matrix~$\bY$ is of rank ${K\!\ll\!M,N}$ so
that we may write ${\bY=\bU\bV^\trans}$, where~$\bU$ and~$\bV$ are
${M\stimes K}$ and ${N\stimes K}$ matrices, respectively.  A common
approach is to say that the rows of~$\bU$ and~$\bV$ are independent draws
from two~$K$-dimensional Gaussian distributions
\citep{salakhutdinov-mnih-2008a}.  We then interpret these~$K$-dimensional
vectors as latent features that are distributed representations of the
members of~$\mcM$ and~$\mcN$.  We denote these vectors as~$\bu_m$
and~$\bv_n$ for the (transposed)~$m$th row of~$\bU$ and~$n$th row of~$\bV$,
respectively, so that~${Y_{m,n}=\bu_m^\trans\bv_n}$.

The distribution linking~$\bY$ and~$\bZ$ is application-specific.  For
ratings data it may be natural to use an ordinal regression model.  For
binary data, such as in link prediction, a Bernoulli logistic model may be
appropriate.  PMF models typically assume that the entries of~$\bZ$ are
independent given~$\bY$, although this is not necessary.  In
Section~\ref{sec:basketball} we will use a conditional likelihood model
that explicitly includes dependencies.

\begin{figure}[t]
  \centering%
  \subfloat[Standard probabilistic matrix factorization]{%
    \includegraphics[width=\linewidth]{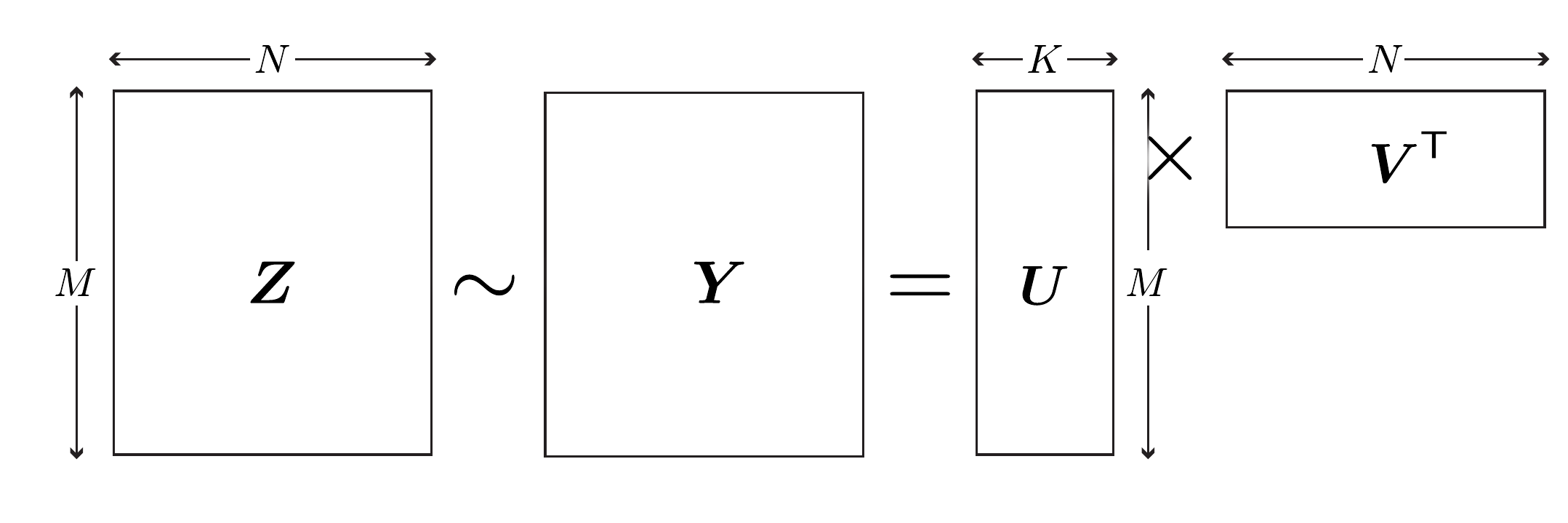}%
  }\\%
  \subfloat[Dependent probabilistic matrix factorization]{%
    \includegraphics[width=\linewidth]{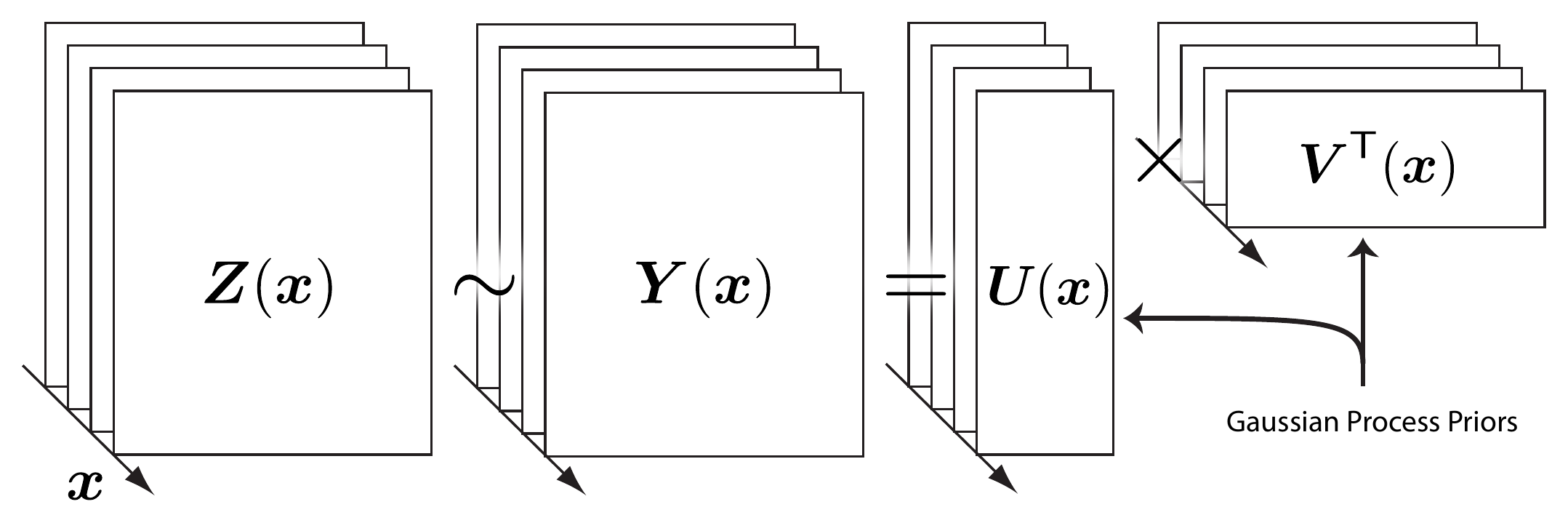}%
  }%
  \caption{\small (a)~The basic low-rank matrix factorization model uses a
    matrix~$\bY$ to parameterize a distribution on random matrices, from
    which~$\bZ$ (containing the observations) is taken to be a sample.  The
    matrix~$\bY$ is the product of two rank-$K$ matrices~$\bU$ and~$\bV$.
    (b)~In dependent probabilistic matrix factorization, we consider the
    low-rank matrices to be ``slices'' of functions over~$\bx$ (coming out
    of the page).  These functions have Gaussian process priors.}%
  \label{fig:feature-funcs}%
\end{figure}

\subsection{Latent Features as Functions}
We now generalize the PMF model to include side information about the
interactions.  Let~$\mcX$ denote the space of such side information,
and~$\bx$ denote a point in~$\mcX$.  The time of a game or of a movie
rating are good examples of such side information, but it could also
involve various features of the interaction, features of the participants,
or general nuisance parameters.

\begin{figure}[t!]
  \centering%
  \subfloat[``Left hand'' latent vector function~$\bu_m(\bx)$]{%
    \includegraphics[width=0.75\linewidth]{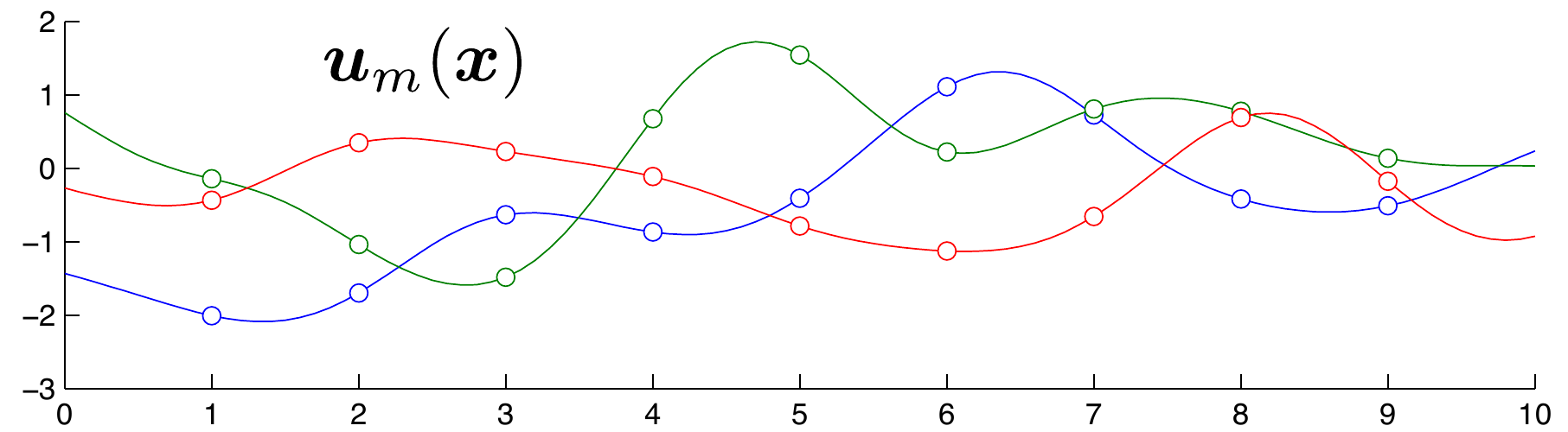}%
  }\\%
  \subfloat[``Right hand'' latent vector function~$\bv_n(\bx)$]{%
    \includegraphics[width=0.75\linewidth]{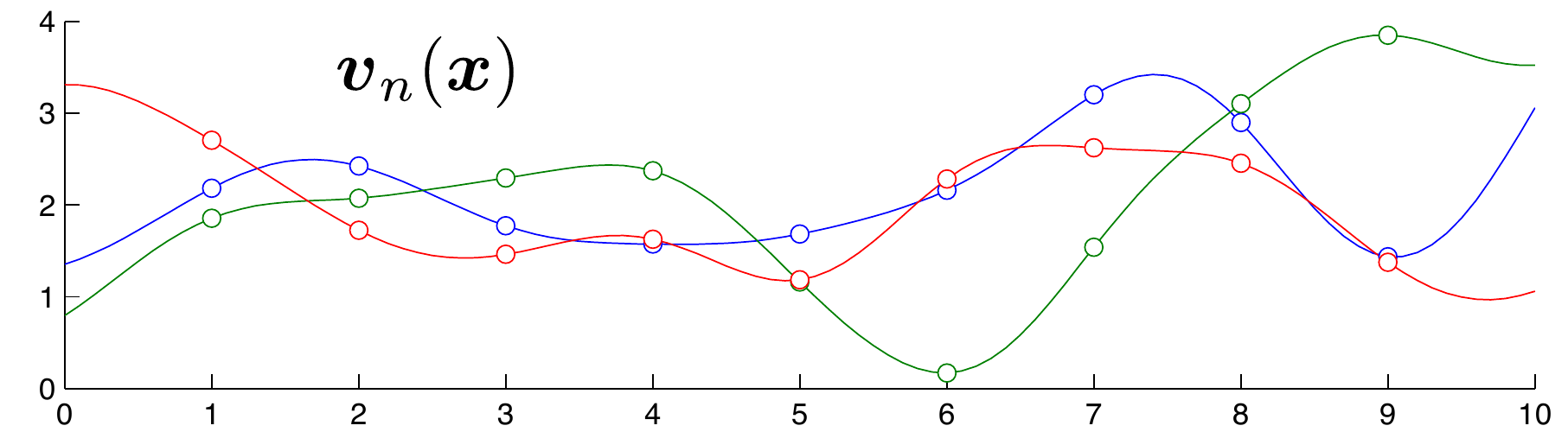}%
  }\\%
  \subfloat[Function of inner product~$Y_{m,n}(\bx)$]{%
    \includegraphics[width=0.75\linewidth]{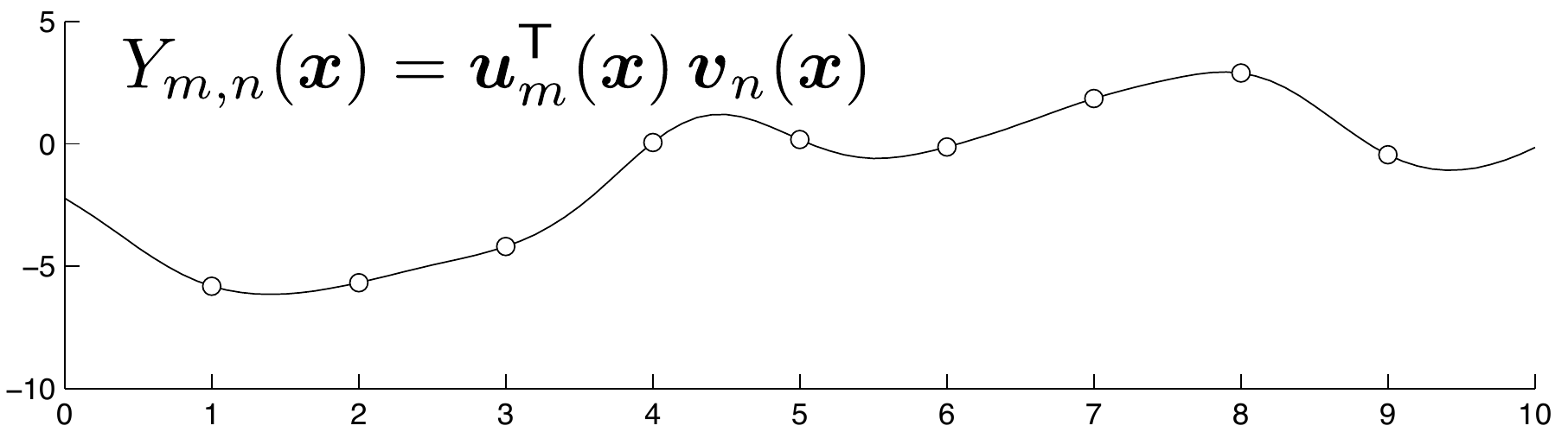}%
  }\\%
  \subfloat[Observed data~$Z_{m,n}$]{%
    \includegraphics[width=0.75\linewidth]{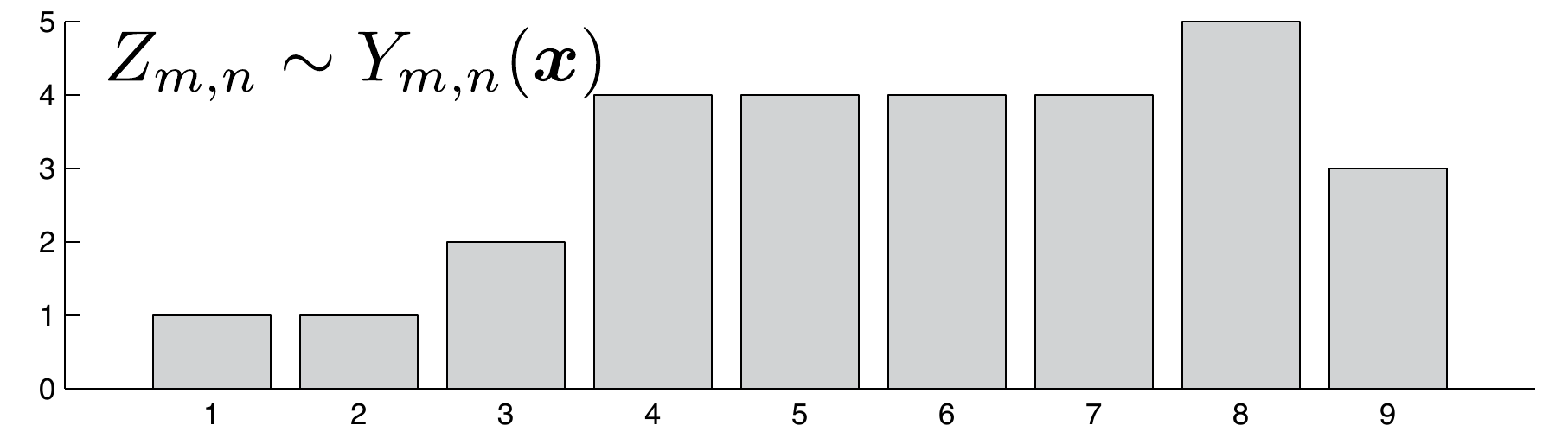}%
  }
  \caption{\small These figures illustrate the generative view of the DPMF
    model. (a,b)~Vector functions for~$m$ and~$n$ are drawn from from
    Gaussian processes.  (c)~The pointwise inner product of these functions
    yields the latent function~$Y_{m,n}(\bx)$.  (d)~The observed data, in
    this case, ordinal values between 1 and 5, that depend
    on~$Y_{m,n}(\bx)$.}%
  \label{fig:func-idea}%
\end{figure}

To enable dependence on such side information, we extend the standard PMF
model by replacing latent feature vectors~$\bu_m$ and~$\bv_n$, with
\textit{latent feature functions} ${\bu_m(\bx):\mcX\to\reals^K}$ and
${\bv_n(\bx):\mcX\to\reals^K}$.  The matrix~$\bY$ is now a
function~$\bY(\bx)$ such that
${Y_{m,n}(\bx)=\bu_m^\trans(\bx)\,\bv_n(\bx)}$, or alternatively,
${\bY(\bx) =\bU(\bx)\bV^\trans(\bx)}$, where the~$\bZ(\bx)$ matrix is drawn
according to a distribution parameterized by~$\bY(\bx)$.  We model
each~$\bZ(\bx)$ conditionally independent, given~$\bY(\bx)$.  This
representation, illustrated in Figures~\ref{fig:feature-funcs}
and~\ref{fig:func-idea}, allows the latent features to vary according
to~$\bx$, capturing the idea that the side information should be relevant
to the distributed representation.  We use a multi-task variant of the
Gaussian process as a prior for these vector functions to construct a
nonparametric Bayesian model of the latent features.

\subsection{Multi-Task Gaussian Process Priors}
When incorporating functions into Bayesian models, we often have general
beliefs about the functions rather than knowledge of a specific basis.  In
these cases, the Gaussian process is a useful prior, allowing for the
general specification of a distribution on functions from~$\mcX$
to~$\reals$ via a positive-definite covariance
kernel~$C(\bx,\bx'):\mcX\times\mcX\to\reals$ and a mean
function~${\mu(\bx):\mcX\to\reals}$.  For a general review of Gaussian
processes for machine learning see \citet{rasmussen-williams-2006a}.  In
this section we restrict our discussion to GP priors for the
functions~$\bu_m(\bx)$, but we deal with the~$\bv_n(\bx)$ functions
identically.

It is reasonable to consider the feature function~$\bu_m(\bx)$ to be
independent of another individual's feature function~$\bu_{m'}(\bx)$,
but we would like for each of the components \textit{within} a
particular function~$\bu_m(\bx)$ to have a structured prior.  Rather
than use independent Gaussian processes for each of the~$K$ scalar
component functions in~$\bu_m(\bx)$, we use a multi-task GP approach
in the vein of \citet{teh-etal-2005a} and \citet{bonilla-etal-2008a}.
We perform a pointwise linear transformation of~$K$ independent latent
functions using a matrix~$L_{\Sigma_U}$ that is the Cholesky
decomposition of an inter-task covariance matrix~$\Sigma_U$, i.e.,
${\Sigma_U=L_{\Sigma_U}L_{\Sigma_U}^\trans}$.  The covariance
functions~$C^U_k(\bx,\bx')$ (and relevant
hyperparameters~$\theta^U_k$) are shared across the members of~$\mcM$,
and constant mean functions~$\bmu_U(\bx)$ are added to each of the
functions after the linear transformation.

The \textit{intra-feature} sharing of the covariance function, mean
function, and hyperparameters is intended to capture the idea that the
characteristic variations of features will tend to be consistent across
members of the set.  If a feature function learns to capture, for example,
whether or not users in a collaborative filtering problem enjoy Christmas
movies, then we might expect them to share an annual periodic variation.
The \textit{inter-feature} covariance matrix~$\Sigma_U$, on the other hand,
captures the idea that some features are informative about others and that
this information can be shared.  \citet{salakhutdinov-mnih-2008a} applied
this idea to scalar features.

We refer to this model as \textit{dependent} probabilistic matrix
factorization, because it ties together a set of PMF problems which
are indexed by~$\mcX$.  This yields a useful spectrum of possible
behaviors: as the length scales of the GP become large and the side
information in~$\bx$ becomes uninformative, then our approach reduces
to a single PMF problem; as the length scales become small and the
side information becomes highly informative then each unique~$\bx$ has
its own PMF model.  The marginal distribution of a PMF problem for a
given~$\bx$ are the same as that in \citet{salakhutdinov-mnih-2008a}.
Additionally, by having each of the~$K$ feature functions use
different hyperparameters, the variation over~$\mcX$ can be shared
differently for the features.  In a sports modeling application, one
feature might correspond to coaches and others to players.  Player
personnel may change at different timescales than coaches and this can
be captured in the model.

\subsection{Constructing Correlation Functions}
\label{sec:corrfuncs}
One of the appeals of using a fully-Bayesian approach is that it in
principle allows hierarchical inference of hyperparameters.  In the DPMF
case, we may not have a strong preconception as to precisely how useful the
side information is for prediction.  The relevance of side information can
be captured by the length-scales of the covariance functions on~$\mcX$
\citep{rasmussen-williams-2006a}.  If~$\mcX$ is a $D$-dimensional real
space, then a standard choice is the automatic relevance determination
(ARD) covariance function:
\begin{align}
  C_{\sf{ARD}}(\bx,\bx') &=
  \exp\left\{-\frac{1}{2}\sum^D_{d=1}(x_d-x'_d)^2/\ell_d^2\right\},
\end{align}
where in our notation there would be~$2K$ sets of length scales that
correspond to the covariance hyperparameters~$\{\theta^U_k\}$
and~$\{\theta^V_k\}$.

While the ARD prior is a popular choice, some DPMF applications may have
temporal data that cause feature functions to fluctuate periodically, as in
the previous Christmas movie example.  For this situation it may be
appropriate to include a periodic kernel such as
\begin{align}
C_{\sf{per}}(x,x') &= \exp\left\{-2\sin^2\left(\frac{1}{2}(x-x')\right)/\ell^2\right\}.
\end{align}
Note that we have defined both of these as \textit{correlation} functions
(unit marginal variances), and allow for variation in function amplitudes
to be captured via~$\Sigma_U$ and~$\Sigma_V$ as in
\citet{bonilla-etal-2008a}.

\subsection{Reducing Multimodality}
\label{sec:multimodality}
When performing inference, overcompleteness can cause difficulties by
introducing spurious modes into the posterior distribution.  It is
useful to construct parameterizations that avoid multimodality.
Although we have developed our notation as~$\bU\bV^\trans$, in
practice we restrict the right-hand factor to be positive via a
component-wise transformation~${\psi(r)\!=\!\ln(1 + e^r)}$ of~$\bV$ so
that ${\bY(\bx)\!=\!\bU(\bx)\psi(\bV^\trans(\bx))}$.  In product
models such as this, there are many posterior modes corresponding to
sign flips in the functions \citep{adams-stegle-2008a}.  Restricting
the sign of one factor improves inference without making the model
less expressive.

\subsection{Summary of Model}
For clarity, we present an end-to-end generative view of the DPMF
model: 1)~Two sets of~$K$ GP hyperparameters, denoted~$\{\theta^U_k\}$
and~$\{\theta^V_k\}$, come from top-hat priors; 2)~$K(M+N)$ functions
are drawn from the~$2K$ Gaussian processes, these are denoted below
as~$\{\bff^U_{k,m}\}$ and~$\{\bff^V_{k,n}\}$; 3)~Two~$K$-dimensional
mean vectors~$\bmu_U$ and~$\bmu_V$ come from vague Gaussian priors;
4)~Two~$K\stimes K$ cross-covariance matrices~$\Sigma_U$
and~$\Sigma_V$ are drawn from uninformative priors on positive
definite matrices; 5)~The ``horizontally sliced''
functions~$\{\bff^U_m\}$ and~$\{\bff^V_n\}$ are transformed with the
Cholesky decomposition of the appropriate cross-covariance matrix and
the mean vectors are added.  6)~The transformation~$\psi(\cdot)$ is
applied elementwise to the resulting~$\{\bv_n(\bx)\}$ to make them
strictly positive; 7)~The inner product of~$\bu_m(\bx)$
and~$\psi(\bv_n(\bx))$ computes~$y_{m,n}(\bx)$; 8)~The
matrix~$\bY(\bx)$ parameterizes a model for the entries of~$\bZ(\bx)$.
Ignoring the vague priors, this is given by:
\begin{align*}
  \bZ(\bx) &\sim p(\bZ\given\bY(\bx)) &
  \bY(\bx) &= \bU(\bx)\,\psi(\bV^\trans(\bx))\\
  \bu_m(\bx) &= L_{\Sigma_U}\bff^U_m + \bmu_U &
  \bv_n(\bx) &= L_{\Sigma_V}\bff^V_n + \bmu_V\\
  \bff^U_{k,m} &\sim \distGP(\bx_m, \theta_k^U) &
  \bff^V_{k,n} &\sim \distGP(\bx_n, \theta_k^V).
\end{align*}

\subsection{Related Models}
\label{sec:related}
There have been several proposals for incorporating side information into
probabilistic matrix factorization models, some of which have used
Gaussian processes.

The Gaussian process latent variable model (GPLVM) is a nonlinear
dimensionality reduction method that can be viewed as a kernelized
version of PCA.  \citet{lawrence-urtasun-2009a} observes that PMF can
be viewed as a particular case of PCA and use the GPLVM as a ``kernel
trick'' on the inner products that produce~$\bY$ from~$\bU\bV^\trans$.
The latent representations are optimized with stochastic gradient
descent.  This model differs from ours in that we us the GP to map
from observed side information to the latent space, while theirs maps
from the latent space into the matrix
entries. \citet{lawrence-urtasun-2009a} also mentions the use of
movie-specific metadata to augment the latent space in their
collaborative filtering application.  We additionally note that our
model allows arbitrary link functions between the latent matrix~$\bY$
and the observations~$\bZ$, including dependent distributions, as
discussed in Section~\ref{sec:bball-specific}.

Another closely-related factorization model is the stochastic
relational model (SRM) \citep{yu-etal-2007a}.  Rather than
representing~$\mcM$ and~$\mcN$ as finite sets, the SRM uses arbitrary
spaces as index sets.  The GP provides a distribution over maps from
this ``identity space'' to the latent feature space.  The SRM differs
from the DPMF in that the input space for our Gaussian process
corresponds to the observations themselves, and not just to the
participants in the relation.  Additionally, we allow each member
of~$\mcM$ and~$\mcN$ to have~$K$ functions, each with a different GP
prior that may have different dependencies on~$\mcX$.

A potential advantage of the DPMF model we present here, relative to
the GPLVM and SRM, is that the GP priors need only be defined on the
data associated with the observations for a single individual.  As
inference in Gaussian processes has cubic computational complexity, it
is preferable to have more independent GPs with fewer data in each one
than a few GPs that are each defined on many thousands of input
points.

There has also been work on explicitly incorporating temporal
information into the collaborative filtering problem, most notably by
the winner of the Netflix prize.  \citet{koren-2009a} included a
simple drift model for the latent user features and baseline ratings.
When rolled into the SVD learning method, this temporal information
significantly improved predictive accuracy.

\section{MCMC Inference and Prediction}
As discussed in Section~\ref{sec:pmf}, the typical objective when
using probabilistic matrix factorization is to predict unobserved
entries in the matrix.  For the DPMF model, as in the Bayesian PMF
model \citep{salakhutdinov-mnih-2008b}, inference and prediction are
not possible in closed form.  We can use Markov chain Monte Carlo
(MCMC), however, to sample from the posterior distribution of the
various parameters and latent variables in the model.  We can then use
these samples to construct a Monte Carlo estimate of the predictive
distribution.  If the entries of interest in~$\bZ(\bx)$ can be easily
sampled given~$\bY(\bx)$ --- as is typically the case --- then samples
from the posterior on~$\bY(\bx)$ allow us to straightforwardly
generate predictive samples, which are the
quantities of interest, and integrate out all of the latent variables.

In the DPMF model, we define the state of the Markov chain with:
1)~the values of the latent feature functions~$\bU(\bx)$
and~$\bV(\bx)$, evaluated at the observations; 2)~the
hyperparameters~$\{\theta^U_k,\theta^V_k\}^K_{k=1}$ associated with
the feature-wise covariance functions, typically capturing the
relevance of the side information to the latent features; 3)~the
feature cross-covariances~$\Sigma_U$ and~$\Sigma_V$; 4)~the feature
function means~$\bmu_U$ and~$\bmu_V$; 5)~any parameters controlling
the conditional likelihood of~$\bZ(\bx)$ given~$\bY(\bx)$.  Note that
due to the convenient marginalization properties of the Gaussian
process, it is only necessary to represent the values of feature
functions at places (in the space of side information) where there
have been observations.

\subsection{Slice Sampling}
When performing approximate inference via Markov chain Monte Carlo,
one constructs a transition operator on the state space that leaves
the posterior distribution invariant.  The transition operator is used
to simulate a Markov chain. Under mild conditions,
the distribution over resulting states
evolves to be closer and closer to the true posterior distribution
(e.g., \citet{neal1993}).
While a generic operator, such as Metropolis--Hastings or Hamiltonian
Monte Carlo, can be implemented, we seek efficient methods that do not
require extensive tuning.  To that end, we use MCMC methods based on
\textit{slice sampling} \citep{neal2003a} when performing inference in
the DPMF model. Some of the variables and parameters required special
treatment, detailed in the next two subsections, for slice sampling to
work well.

\subsection{Elliptical Slice Sampling}
Sampling from the posterior distribution over latent functions with
Gaussian process priors is often a difficult task and can be slow to
mix, due to the structure imposed by the GP prior.  In this case, we
have several collections of functions in ~$\bU(\bx)$ and~$\bV(\bx)$
that do not lend themselves easily to typical methods such as Gibbs
sampling.  Recently, a method has been developed to specifically
enable efficient slice sampling of complicated Gaussian process models
with no tuning or gradients \citep{murray-adams-mackay-2010a}.  This
method, called \textit{elliptical slice sampling} (ESS), takes
advantage of invariances in the Gaussian distribution to make
transitions that are never vetoed by the highly-structured GP
prior, even when there are a large number of such functions as in the
DPMF.

\subsection{Sampling GP Hyperparameters}
\label{sec:gphypers}
As discussed in Section~\ref{sec:corrfuncs}, the length scales in the
covariance (correlation) functions of the Gaussian processes play a
critical role in the DPMF model.  It is through these hyperparameters that
the model weighs the effect of side information on the predictions.  In the
DPMF model, a set of hyperparameters~$\theta^U_k$ (or~$\theta^V_k$)
affect~$M$ (or~$N$) functions.  The typical approach to this would be to
fix the relevant functions~$\{\bff^U_{k,m}\}^M_{m=1}$ and sample
from the conditional posterior:
\begin{align*}
  p(\theta^U_k\given \{\bff^U_{k,m}\}^M_{m=1}) &\propto
  p(\theta^U_k)\prod^M_{m=1}\distNorm(\bff^U_{k,m};\, 0, \Xi^U_{k,m}),
\end{align*}
where~$\Xi^U_{k,m}$ is the matrix that results from applying the
covariance function with hyperparameters~$\theta^U_k$ to the set of
side information for~$m$.  In practice, the Markov chain on this
distribution can mix very slowly, due to the strong constraints
arising from the~$M$ functions, despite the relative weakness of the
conditional likelihood on the data. Therefore, we use an approach
similar to \citep{christensen2002}, which mixes faster in our
application, but still leaves the posterior distribution on the
hyperparameters invariant.

The model contains several draws from a Gaussian process of the form:
$\bff^V_{k,n} \!\sim\! \distGP(\bx_n, \theta_k^V)$. Consider a vector of
evaluations of one of these latent functions that is marginally distributed as
$\bff\!\sim\!\distNorm(\bmm,\Xi_\theta)$. Under the generative process, the
distribution over the latent values is strongly dependent on the hyperparameters
$\theta$ that specify the covariance. As a result, the posterior conditional
distribution over the hyperparameters for fixed latent values will be strongly
peaked, leading to slow mixing of a Markov chain that updates $\theta$ for
fixed~$\bff$.
Several authors have found it useful to reparameterize Gaussian models so that
under the prior the latent values are independent of each other and the
hyperparameters.
This can be achieved by setting $\bnu\!=\!L_\theta^{-1} (\bff\!-\!\bmm)$,
where $L_\theta$ is a matrix square root, such as the Cholesky decomposition, of
the covariance~$\Xi_\theta$. Under the new prior representation, $\bnu$ is
drawn from a spherical unit Gaussian for all~$\theta$.

We slice sample the GP hyperparameters after reparameterizing all vectors of
latent function evaluations. As the hyperparameters change, the function values
$\bff\!=\!\bmm\!+\!L_\theta\bnu$ will also change to satisfy the covariance structure of
the new settings. Having observed data, some~$\bff$ settings are very unlikely;
in the reparameterized model the likelihood terms will restrict how much the
hyperparmaters can change. In the application we consider, with very
noisy data, these updates work much better than updating the
hyperparameters for fixed~$\bff$.
In problems where the data strongly restrict the possible changes in
$\bff$, more advanced reparameterizations are possible
\citep{christensen2006}. We have developed related slice sampling
methods that are easy to apply \citep{murray-adams-2010a}.

\section{DPMF for Basketball Outcomes}
\label{sec:basketball}
To demonstrate the utility of the DPMF approach, we use our method to
model the scores of games in the National Basketball Association (NBA)
in the years 2002 to 2009.  This task is an appealing one to study for
several reasons: 1)~it is of a medium size, with approximately ten
thousand observations; 2)~it provides a natural censored-data
evaluation setup via a ``rolling predictions'' problem; 3)~expert
human predictions are available via betting lines; 4)~the properties
of teams vary over time as players are traded, retire and are injured;
5)~other side information, such as which team is playing at home, is
clearly relevant to game outcomes.  For these reasons, using
basketball as a testbed for probabilistic models is not a new idea.
In the statistics literature there have been previous studies of
collegiate basketball outcomes by \citet{Schwertman-1991-PMN},
\citet{Schwertman-1996-MPM}, and \cite{Carlin-1996-INB}, although with
smaller data sets and narrower models.

We use the DPMF to model the scores of games. The observations in the
matrix~$\bZ(\bx)$ are the actual scores of the games with side
information~$\bx$.  $Z_{m,n}(\bx)$~is the number of points scored by
team~$m$ against team~$n$, and~$Z_{n,m}(\bx)$ is the number of points
scored by team~$n$ against team~$m$.  We model these with a bivariate
Gaussian distribution, making this a somewhat unusual PMF-type model
in that we see \textit{two} matrix entries with each observation and
we place a joint distribution over them.  We use a single variance for
all observations and allow for a correlation between the scores of the
two teams.  While the Gaussian model is not a perfect match for the
data --- scores are non-negative integers -- each team in the NBA
tends to score about 100 points per game, with a standard deviation of
about ten so that very little mass is assigned to negative numbers.

Even though both sets in this dyadic problem are the same,
i.e.,~$\mcM\!=\!\mcN$, we use different latent feature functions
for~$\bU(\bx)$ and~$\bV(\bx)$.  This makes the~$\bU(\bx)$ functions of
offense and the~$\bV(\bx)$ functions of defense, since one of them
contributes to ``points for'' and the other contributes to ``points
against''.  This specialization allows the Gaussian process
hyperparameters to have different values for offense and defense,
enabling the side information to modulate the number of points scored
and conceded in potentially different ways.

\subsection{Problem Setup}
To use NBA basketball score prediction as a task to determine the
value that using side information in our framework provides relative
to the standard PMF model, we set up a rolling censored-data problem.
We divided each of the eight seasons into four-week blocks.  For each
four-week block, the models were asked to make predictions about the
games during that interval using only information from the past.  In
other words, when making predictions for the month of February 2005,
the model could only train on data from 2002 through January 2005.  We
rolled over each of these intervals over the entire data set,
retraining the model each time and evaluating the predictions.  We
used three metrics for evaluation: 1)~mean predictive log probability
from a Rao--Blackwellized estimator; 2)~error in the binary
winner-prediction task; 3)~root mean-squared error (RMSE) of the
two-dimensional score vector.

The winner accuracies and RMSE can be compared against human experts,
as determined by the betting lines associated with the games.  Sports
bookmakers assign in advance two numbers to each game, the
\textit{spread} and the \textit{over/under}.  The spread is a number
that is added to the score of a specified team to yield a bettor an
even-odds return.  For example, if the spread between the LA Lakers
and the Cleveland Cavaliers is ``-4.5 for the Lakers'' then a
single-unit bet for the Lakers yields a single-unit return if the
Lakers win by 4.5 points or more (the half-point prevents ties, or
\textit{pushes}).  If the Lakers lose or beat the Cavaliers by fewer
than 4.5 points, then a single-unit bet on the Cavaliers would win a
single-unit return.  The over/under determines the threshold of a
single-unit bet on the sum of the two scores.  For example, if the
over/under is 210.5 and the final score is 108 to 105, then a bettor
who ``took the over'' with a single-unit bet would win a single-unit
return, while a score of 99 to 103 would cause a loss (or a win to a
bettor who ``took the under'').

From the point of view of model evaluation, these are excellent
predictions, as the spread and over/under themselves are set by the
bookmakers to balance bets on each side.  This means that expert
humans exploit any data available (e.g., referee identities and injury
reports, which are not available to our model) to exert market forces
that refine the lines to high accuracy.  The sign of the spread
indicates the favorite to win.  To determine the implied score
predictions themselves, we can solve a simple linear system:
\begin{align}
  \begin{bmatrix}
    1 & 1\\
    1 & -1
  \end{bmatrix}
  \begin{bmatrix}
    \text{away score}\\
    \text{home score}
  \end{bmatrix}
  &=
  \begin{bmatrix}
    \text{over/under}\\
    \text{home spread}
  \end{bmatrix}.
\end{align}

\subsection{Basketball-Specific Model Aspects}
\label{sec:bball-specific}
As mentioned previously, the conditional likelihood function that
parameterizes the distribution over the entries in~$\bZ(\bx)$ in terms
of~$\bY(\bx)$ is problem specific.  In this application, we use
\begin{align}
  \label{eqn:score-dist}
  \begin{bmatrix}
    Z_{m,n}(\bx)\\
    Z_{n,m}(\bx)
  \end{bmatrix}
  &\sim
  \distNorm\left(
    \begin{bmatrix}
      Y_{m,n}(\bx)\\
      Y_{n,m}(\bx)
    \end{bmatrix},
    \begin{bmatrix}
      \sigma^2 & \rho\sigma^2\\
      \rho\sigma^2 & \sigma^2
      \end{bmatrix}
    \right),
\end{align}
where~${\sigma\in\reals^{+}}$ and~${\rho\in(-1,1)}$ parameterize the
bivariate Gaussian on scores and are included as part of inference.
(A typical value for the correlation coefficient
was~${\rho\!=\!0.4}$.)  This allows us to easily compute the
predictive log probabilities of the censored test data using a
Rao--Blackwellized estimator.  To do this, we sample and store
predictive state from the Markov chain to construct a Gaussian mixture
model.  Given the predictive samples of the latent function at the new
time, we can compute the means for the distribution in
Equation~\ref{eqn:score-dist}.  The covariance parameters are also
being sampled and this forms one component in a mixture model with
equal component weights.  Over many samples, we form a good predictive
estimate.

\subsection{Nonstationary Covariance}
\begin{figure}[t!]
  \centering%
  \includegraphics[width=0.75\linewidth]{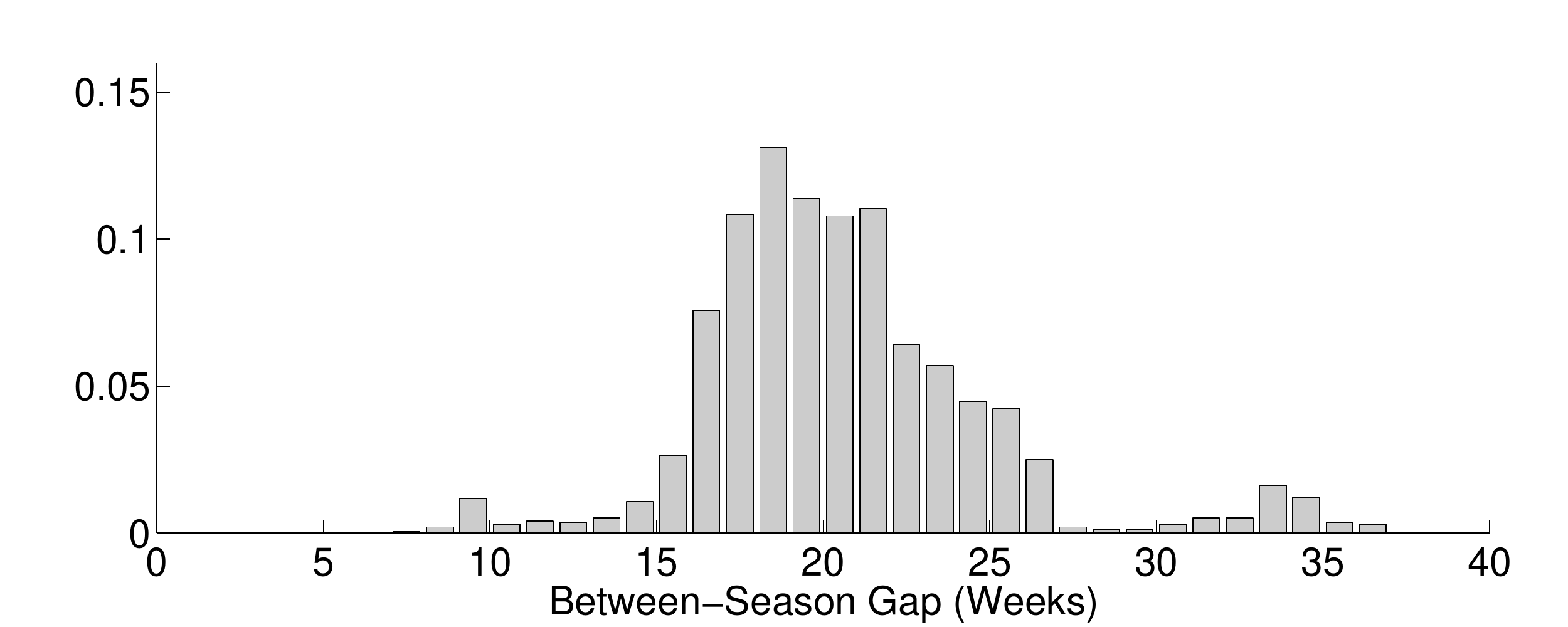}%
  \caption{\small Histogram of the gap between seasons.}%
  \label{fig:gaps}%
\end{figure}

In the DPMF models incorporating temporal information, we are
attempting to capture fluctuations in the latent features due to
personnel changes, etc.  One unusual aspect of this particular
application is that we expect the notion of time scale to vary
depending on whether it is the off-season.  The timescale appropriate
during the season is almost certainly inappropriate to describe the
variation during the 28 weeks between the end of one regular season and
the start of another.  To handle this nonstationarity of the data, we
introduced an additional parameter that is the effective number of
weeks between seasons, which we expect to be smaller than the true
number of weeks.  We include this as a hyperparameter in the
covariance functions and include it as part of inference, using the
same slice sampling technique described in Section~\ref{sec:gphypers}.
A histogram of inferred gaps for~$K\!=\!4$ is shown in
Figure~\ref{fig:gaps}.  Note that most of the mass is below the true
number of weeks.

\subsection{Experimental Setup and Results}
We compared several different model variants to evaluate the utility
of side information.  We implemented the standard fully-Bayesian PMF
model using the same likelihood as above, generating predictive log
probabilities using the same method as for the DPMF.  We constructed
DPMFs with temporal information, binary home/away information, and
both of these together.  We applied each of these models using
different numbers of latent features, $K$, from one to five.  We ran
ten separate Markov chains to predict each censored interval.  Within
a single year, we initialized the Markov state from the ending state
of the previous chain, for a ``warm start''.  The ``cold start'' at
the beginning of the year ran for 1000 burnin iterations, while warm
starts ran for 100 iterations in each of the ten chains.  After
burning in and thinning by a factor of four, 100 samples of each
predictive score were kept from each chain, resulting in 1000
components in the predictive Gaussian mixture model.

To prevent the standard PMF model from being too heavily influenced by
older data, we only provided data to it from the current season and
the previous two seasons.  To prevent an advantage for the DPMF, we
also limited its data in the same way.  Sampling from the covariance
hyperparameters in the model is relatively expensive, due to the need
to compute multiple Cholesky decompositions.  To improve efficiency in
this regard, we ran an extensive Markov chain to burn in these
hyperparameters and then fixed them for all further sampling.  Without
hyperparameter sampling, the remainder of the MCMC state can be
iterated in approximately three minutes on a single core of a modern
workstation.  We performed this burn-in of hyperparameters on a span of
games from 2002 to 2004 and ultimately used the learned parameters for
prediction, so there is a mild amount of ``cheating'' on 2004 and
before as those data have technically been seen already.  We believe
this effect is very small, however, as the covariance hyperparameters
(the only state carried over) are only loosely connected to the data.

\begin{figure}[t!]
  \centering%
  \subfloat[{\scriptsize LA at home, week 10, 2004}]{%
    \includegraphics[width=0.24\linewidth]
    {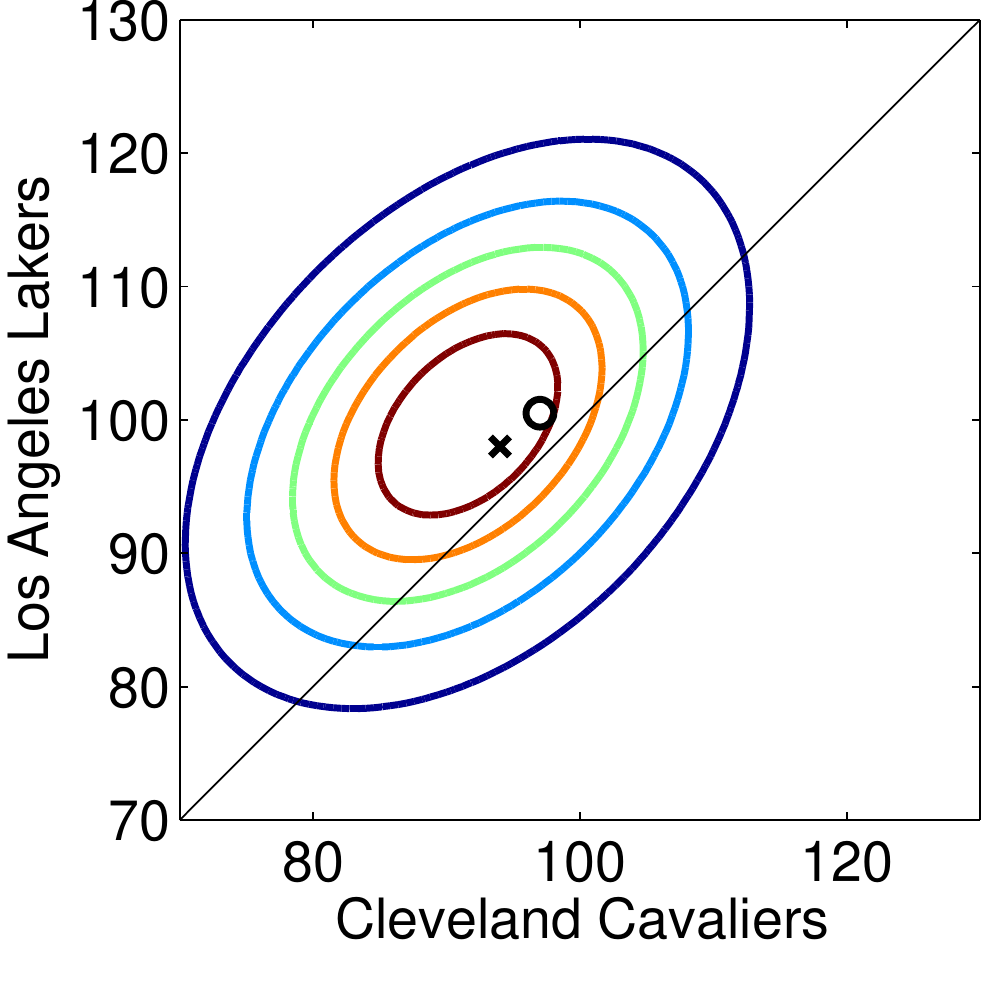}%
  }~%
  \subfloat[{\scriptsize Cle at home, week 14, 2004}]{%
    \includegraphics[width=0.24\linewidth]
    {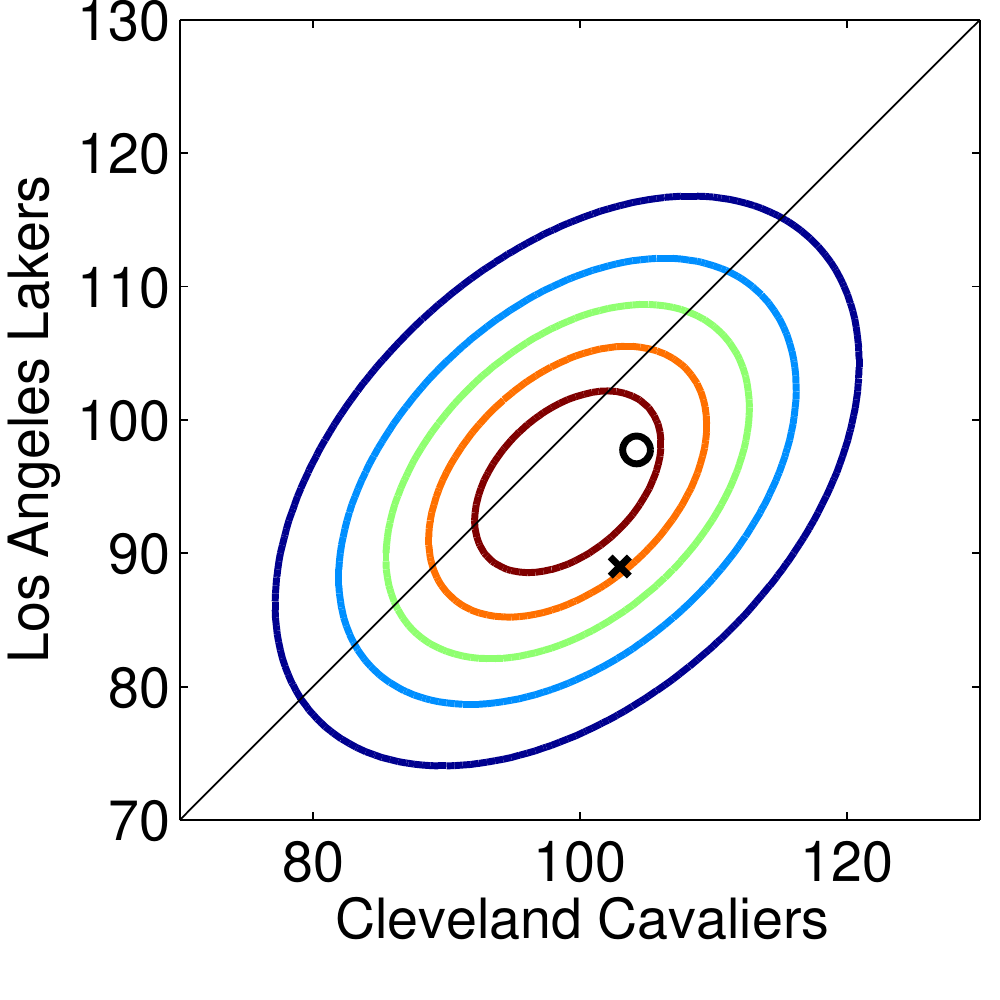}%
  }~%
  \subfloat[{\scriptsize LA at home, week 10, 2008}]{%
    \includegraphics[width=0.24\linewidth]
    {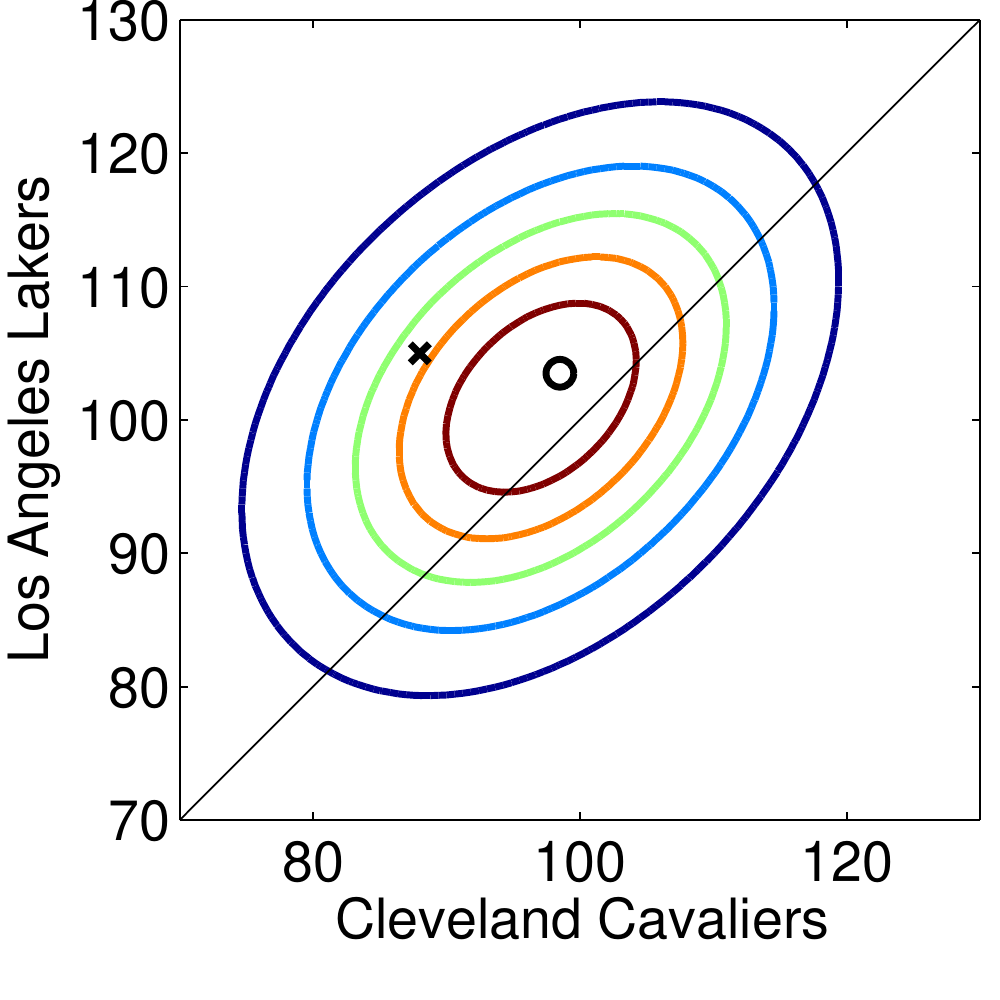}%
  }~%
  \subfloat[{\scriptsize Cle at home, week 12, 2008}]{%
    \includegraphics[width=0.24\linewidth]
    {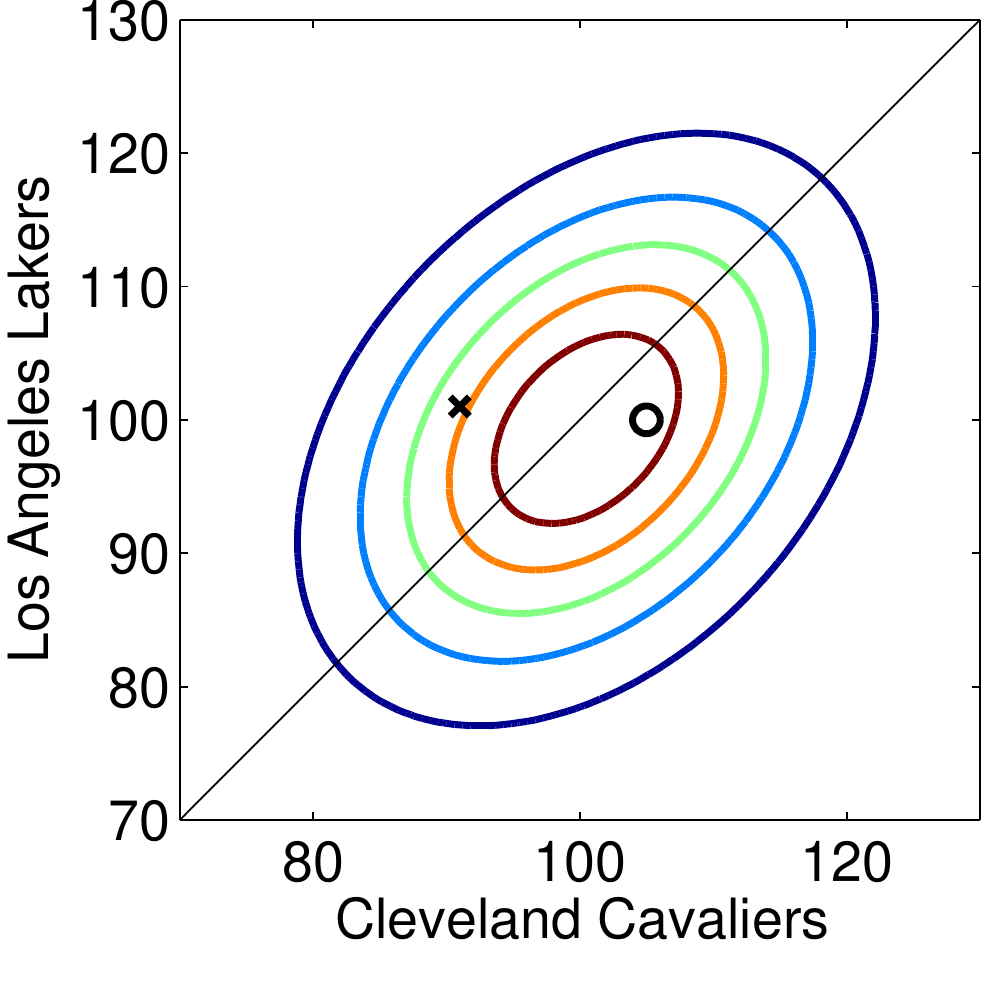}%
  }%
  \caption{\small Contour plots showing the predictive densities for
    four games between the Cleveland Cavaliers and the Los Angeles Lakers,
    using~$K=3$ with home/away and temporal information available.  Los
    Angeles is the home team in~(a) and~(c). Cleveland is the home team
    in~(b) and~(d).  Figures~(a) and~(b) were in the 2004 season,~(c)
    and~(d) were in the 2008 season.  The ``\textsf{o}'' shows the
    expert-predicted score and the ``\textsf{x}'' shows the true outcome.
    The diagonal line indicates the winner threshold.  Note the substantial
    differences between home and away, even when the times are close to
    each other.}
  \label{fig:scores}
\end{figure}

Results from these evaluations are provided in
Table~\ref{tbl:results}.  The DPMF model demonstrated notable
improvements over the baseline Bayesian PMF model, and the inclusion
of more information improved predictions over the time and home/away
information alone.  The predictions in 2009 are less consistent, which
we attribute to variance in evaluation estimates from fewer intervals
being available as the season is still in progress.  The effect of the
number of latent features~$K$ in the complex model is much less clear.
Figure~\ref{fig:scores} shows the joint predictions for four different
games between the Cleveland Cavaliers and the Los Angeles Lakers,
using~$K=3$ with time and home/away available.  The differences
between the predictions illustrate that the model is incorporating
information both from home advantage and variation over time.

\section{Discussion}
In this paper we have presented a nonparametric Bayesian variant of
probabilistic matrix factorization that induces dependencies between
observations via side information using Gaussian processes.  This model has
the convenient property that, conditioned on the side information, the
marginal distributions are equivalent to those in well-studied existing PMF
models.  While Gaussian processes and MCMC often carry a signficant
computational cost, we have developed a framework that can make useful
predictions on real problems in a practical amount of time --- hours for
most of the predictions in the basketball problem we have studied.

There are several interesting ways in which this work could be extended.
One notable issue that we have overlooked and would be relevant for many
applications is that the Gaussian processes as specified in
Section~\ref{sec:corrfuncs} only allow for smooth variation in latent
features.  This slow variation may be inappropriate for many models: if a
star NBA player has a season-ending injury, we would expect that to be
reflected better in something like a changepoint model (see, e.g.,
\citet{barry-hartigan-1993a}) than a GP model.  Also, we have not addressed
the issue of how to select the number of latent features,~$K$, or how to
sample from this parameter as part of the model.  Nonparametric Bayesian
models such as those propsed by \citet{meeds-etal-2006a} may give insight
into this problem.  Finally, we should note that other authors have
explored other kinds of structured latent factors (e.g.,
\citet{sutskever-etal-2009a}), and there may be interesting ways to combine
the features of these approaches with the DPMF.

\begin{table*}[t!]
  \centering%
  \caption{\small Evaluations of PMF and DPMF algorithms with various numbers of
    latent factors.  The PMF model is the fully-Bayesian approach of
    \citet{salakhutdinov-mnih-2008b}, with our application-specific
    likelihood.  DPMF(t) is the DPMF with only temporal information,
    DPMF(h) has only binary home/away indicators, DPMF(t,h) has both
    temporal and home/away information.  (a)~Mean predictive log
    probabilities of test data.  (b)~Error rates of winner prediction are
    on the left, RMSEs of scores are on the right.  Expert human
    predictions are shown on the bottom.}
  \subfloat[Mean log probabilities for rolling score prediction]{%
    \setlength{\tabcolsep}{4pt}%
    \resizebox{\textwidth}{!}{%
      \begin{tabular}{c c r r r r r r r r r}
 & &  2002 & 2003 & 2004 & 2005 & 2006 & 2007 & 2008 & 2009  & All \\
\hline \multirow{5}{*}{\small{PMF}}  &  $K1$  &  -7.644 & -7.587 & -7.649 & -7.580 & -7.699 & -7.733 & -7.634 & -7.653 & -7.647  \\
&  $K2$  &  -7.623 & -7.587 & -7.654 & -7.581 & -7.704 & -7.738 & -7.638 & -7.673 & -7.650  \\
&  $K3$  &  -7.615 & -7.586 & -7.652 & -7.581 & -7.698 & -7.734 & -7.637 & -7.666 & -7.646  \\
&  $K4$  &  -7.619 & -7.585 & -7.653 & -7.581 & -7.703 & -7.734 & -7.635 & -7.667 & -7.647  \\
&  $K5$  &  -7.641 & -7.589 & -7.653 & -7.580 & -7.700 & -7.736 & -7.638 & -7.667 & -7.650  \\
\hline \multirow{5}{*}{\small{DPMF(t)}}  &  $K1$  &  -7.652 & -7.535 & -7.564 & -7.559 & -7.660 & -7.665 & -7.618 & -7.703 & -7.620  \\
&  $K2$  &  -7.620 & -7.551 & -7.580 & -7.544 & -7.675 & -7.658 & -7.611 & -7.724 & -7.621  \\
&  $K3$  &  -7.620 & -7.560 & -7.605 & -7.549 & -7.673 & -7.669 & -7.611 & -7.635 & -7.615  \\
&  $K4$  &  -7.618 & -7.549 & -7.585 & -7.548 & -7.673 & -7.670 & -7.608 & -7.651 & -7.613  \\
&  $K5$  &  -7.640 & -7.558 & -7.591 & -7.554 & -7.669 & -7.670 & -7.609 & -7.651 & -7.618  \\
\hline \multirow{5}{*}{\small{DPMF(h)}}  &  $K1$  &  -7.639 & -7.549 & -7.624 & -7.549 & -7.670 & -7.706 & -7.606 & -7.627 & -7.621  \\
&  $K2$  &  -7.587 & -7.553 & -7.626 & -7.551 & -7.670 & -7.707 & -7.613 & -7.640 & -7.618  \\
&  $K3$  &  -7.580 & -7.542 & -7.618 & -7.539 & -7.667 & -7.706 & -7.602 & -7.637 & -7.612  \\
&  $K4$  &  -7.587 & -7.545 & -7.623 & -7.547 & -7.673 & -7.704 & -7.612 & -7.652 & -7.618  \\
&  $K5$  &  -7.594 & -7.541 & -7.619 & -7.544 & -7.669 & -7.709 & -7.606 & -7.643 & -7.616  \\
\hline \multirow{5}{*}{\small{DPMF(t,h)}}  &  $K1$  &  -7.656 & -7.515 & -7.562 & -7.534 & -7.659 & -7.662 & -7.602 & -7.670 & -7.607  \\
&  $K2$  &  -7.585 & -7.515 & -7.560 & -7.520 & -7.646 & \textbf{-7.639} & -7.591 & -7.695 & -7.594  \\
&  $K3$  &  \textbf{-7.579} & -7.516 & -7.563 & -7.524 & -7.651 & -7.643 & \textbf{-7.575} & \textbf{-7.586} & \textbf{-7.580}  \\
&  $K4$  &  -7.584 & \textbf{-7.511} & \textbf{-7.546} & -7.520 & -7.640 & -7.643 & -7.582 & -7.620 & -7.581  \\
&  $K5$  &  -7.593 & -7.515 & -7.569 & \textbf{-7.512} & \textbf{-7.634} & -7.640 & -7.589 & -7.637 & -7.586  \\
\hline\end{tabular}
      }%
    }\\%
    \subfloat[Left: percentage error on rolling winner prediction, Right: RMSE on rolling score prediction]{%
      \setlength{\tabcolsep}{1pt}%
      \resizebox{\textwidth}{!}{%
          \begin{tabular}{c c r r r r r r r r r c|r r r r r r r r r}
 & &  2002 & 2003 & 2004 & 2005 & 2006 & 2007 & 2008 & 2009  & All & & 2002 & 2003 & 2004 & 2005 & 2006 & 2007 & 2008 & 2009  & All \\
\hline \multirow{5}{*}{\small{PMF}}  &  $K1$  &  38.9 & 39.4 & 41.8 & 37.6 & 41.0 & 38.2 & 36.7 & 36.7 & 38.8  & \rule{0.15cm}{0cm} &  16.66 & 15.92 & 16.80 & 15.84 & 17.16 & 17.39 & 16.38 & 16.86 & 16.63  \\
&  $K2$  &  37.8 & 38.5 & 42.1 & 37.1 & 40.7 & 37.9 & 36.4 & 38.6 & 38.6  & \rule{0.15cm}{0cm} &  16.38 & 15.91 & 16.82 & 15.85 & 17.16 & 17.41 & 16.33 & 16.96 & 16.61  \\
&  $K3$  &  37.2 & 38.7 & 42.4 & 37.0 & 40.5 & 37.5 & 36.8 & 38.1 & 38.5  & \rule{0.15cm}{0cm} &  16.35 & 15.89 & 16.81 & 15.85 & 17.12 & 17.38 & 16.34 & 16.92 & 16.59  \\
&  $K4$  &  37.5 & 38.2 & 41.7 & 36.7 & 40.3 & 37.8 & 36.1 & 37.6 & 38.2  & \rule{0.15cm}{0cm} &  16.34 & 15.90 & 16.81 & 15.84 & 17.15 & 17.39 & 16.34 & 16.93 & 16.59  \\
&  $K5$  &  39.1 & 38.1 & 41.4 & 37.1 & 41.0 & 37.8 & 36.1 & 38.6 & 38.6  & \rule{0.15cm}{0cm} &  16.41 & 15.93 & 16.81 & 15.83 & 17.14 & 17.40 & 16.35 & 16.93 & 16.61  \\
\hline \multirow{5}{*}{\small{DPMF(t)}}  &  $K1$  &  37.8 & 37.7 & 37.9 & 37.3 & 39.1 & 34.4 & 34.4 & 35.7 & 36.8  & \rule{0.15cm}{0cm} &  16.73 & 15.68 & 16.07 & 15.54 & 16.73 & 16.93 & 16.42 & 17.46 & 16.46  \\
&  $K2$  &  37.4 & 38.0 & 37.5 & 39.2 & 40.3 & 34.3 & 33.6 & 37.6 & 37.3  & \rule{0.15cm}{0cm} &  16.37 & 15.70 & 16.30 & 15.29 & 16.84 & 16.75 & 16.21 & 17.57 & 16.39  \\
&  $K3$  &  37.2 & 38.9 & 39.0 & 36.5 & 38.1 & 36.3 & 34.0 & 32.4 & 36.5  & \rule{0.15cm}{0cm} &  16.39 & 15.77 & 16.49 & 15.50 & 16.97 & 16.86 & 16.05 & \textbf{16.67} & 16.34  \\
&  $K4$  &  37.7 & 38.6 & 37.4 & 36.0 & 38.8 & 35.7 & 34.6 & 35.7 & 36.8  & \rule{0.15cm}{0cm} &  16.37 & 15.63 & 16.30 & 15.46 & 16.90 & 16.90 & 16.11 & 16.91 & 16.33  \\
&  $K5$  &  38.6 & 39.4 & 38.0 & 35.9 & 38.4 & 37.0 & 34.1 & 36.7 & 37.2  & \rule{0.15cm}{0cm} &  16.39 & 15.71 & 16.37 & 15.48 & 16.90 & 16.85 & 16.16 & 16.84 & 16.35  \\
\hline \multirow{5}{*}{\small{DPMF(h)}}  &  $K1$  &  37.3 & 37.7 & 39.3 & 34.3 & 38.1 & 37.9 & 34.3 & \textbf{29.5} & 36.1  & \rule{0.15cm}{0cm} &  16.62 & 15.88 & 16.40 & 15.59 & 16.87 & 17.01 & 16.14 & 16.70 & 16.41  \\
&  $K2$  &  36.8 & 37.1 & 38.4 & 34.3 & 38.3 & 37.6 & 34.7 & 31.0 & 36.0  & \rule{0.15cm}{0cm} &  16.11 & 15.93 & 16.44 & 15.60 & 16.88 & 17.04 & 16.15 & 16.79 & 16.37  \\
&  $K3$  &  36.8 & 37.4 & 38.4 & 34.9 & 36.8 & 38.6 & 34.1 & 30.5 & 35.9  & \rule{0.15cm}{0cm} &  15.91 & 15.92 & 16.25 & 15.42 & 16.81 & 16.87 & 16.05 & 16.73 & 16.25  \\
&  $K4$  &  36.6 & 37.8 & 38.4 & 35.1 & 37.9 & 38.1 & 34.6 & 31.4 & 36.2  & \rule{0.15cm}{0cm} &  15.92 & 15.88 & 16.35 & 15.51 & 16.88 & 16.98 & 16.13 & 16.90 & 16.33  \\
&  $K5$  &  \textbf{35.1} & 36.9 & 38.5 & 34.5 & 37.4 & 38.0 & 34.1 & 30.0 & 35.6  & \rule{0.15cm}{0cm} &  16.08 & 15.89 & 16.28 & 15.50 & 16.85 & 17.05 & 16.08 & 16.77 & 16.32  \\
\hline \multirow{5}{*}{\small{DPMF(t,h)}}  &  $K1$  &  37.2 & 36.1 & 36.7 & 35.3 & 38.6 & 33.7 & 32.0 & 37.1 & 35.8  & \rule{0.15cm}{0cm} &  16.76 & \textbf{15.55} & 16.07 & 15.46 & 16.69 & 16.91 & 16.26 & 17.26 & 16.38  \\
&  $K2$  &  36.1 & 37.0 & 36.3 & 35.4 & 38.8 & 33.6 & 32.5 & 36.7 & 35.8  & \rule{0.15cm}{0cm} &  16.08 & 15.58 & 16.04 & \textbf{15.19} & 16.59 & 16.61 & 16.07 & 17.25 & 16.19  \\
&  $K3$  &  37.0 & \textbf{34.9} & 34.4 & 33.8 & 37.0 & 34.2 & 31.7 & 30.0 & 34.1  & \rule{0.15cm}{0cm} &  \textbf{15.90} & 15.72 & 15.89 & 15.35 & 16.69 & \textbf{16.57} & \textbf{15.86} & 16.71 & 16.10  \\
&  $K4$  &  36.0 & 35.2 & 34.1 & 35.0 & 37.4 & \textbf{33.2} & \textbf{31.5} & 30.5 & 34.1  & \rule{0.15cm}{0cm} &  15.93 & 15.62 & \textbf{15.75} & 15.20 & \textbf{16.52} & 16.60 & 16.03 & 16.81 & \textbf{16.07}  \\
&  $K5$  &  35.3 & 35.6 & \textbf{34.0} & \textbf{33.2} & \textbf{36.2} & 33.5 & 31.7 & 32.9 & \textbf{34.0}  & \rule{0.15cm}{0cm} &  16.05 & 15.66 & 15.96 & 15.21 & 16.57 & 16.60 & 16.00 & 16.97 & 16.14  \\
\hline \multicolumn{2}{c}{Expert} &  30.9 & 32.2 & 29.7 & 31.4 & 33.3 & 30.6 & 29.8 & 29.4 & 30.9 & \rule{0.15cm}{0cm} &  14.91 & 14.41 & 14.55 & 14.70 & 15.36 & 15.18 & 14.95 & 15.49 & 14.95 \\
\hline\end{tabular}
        }%
      }
      \label{tbl:results}
\end{table*}
\afterpage{\clearpage}

\subsection*{Acknowledgements}
The authors wish to thank Amit Gruber, Geoffrey Hinton and Rich Zemel
for valuable discussions.  The idea for placing basketball scores
directly into the matrix was originally suggested by Danny Tarlow.
RPA is funded by the Canadian Institute for Advanced Research.

\bibliographystyle{unsrtnat}
\bibliography{draft}

\end{document}